\documentclass{article}

\PassOptionsToPackage{numbers, sort&compress}{natbib}
\usepackage[preprint]{neurips_2026}

\usepackage[utf8]{inputenc}
\usepackage[T1]{fontenc}
\usepackage{hyperref}
\usepackage{url}
\usepackage{booktabs}
\usepackage{placeins}
\usepackage{amsfonts,amsmath,amssymb,amsthm}
\usepackage{microtype}
\usepackage{graphicx}
\usepackage{xcolor}
\hypersetup{colorlinks=true, linkcolor=blue!50!black, citecolor=blue!50!black, urlcolor=blue!50!black}

\newtheorem{lemma}{Lemma}
\newtheorem{proposition}[lemma]{Proposition}
\newcommand{\floatconts}[3]{#2\label{#1}{\centering #3\par}}

\newcommand{\codeavailability}{Code, per-run results and a script that regenerates the tables and checks numerical claims are available at \url{https://github.com/aleetreny/adamw-shortcut-geometry}.}

\title{Spreading a Shortcut Can Reverse AdamW's Advantage over L2}

\author{%
  Alejandro Treny Ortega\\
  University College London\\
  \texttt{alejandro.ortega.26@ucl.ac.uk}
}



\begin{document}

\maketitle

\begin{abstract}
Two shortcuts can carry exactly the same signal and still mean different things to an optimiser. I study a two-layer ReLU network on MNIST with a spurious shortcut that agrees with the label with probability 0.95 and has the same $\ell_2$ norm whether it occupies 4 or 196 extra input coordinates. For SGD with coupled $\ell_2$ regularisation the two cases are the same problem in different orthonormal bases, and with transported initial weights its predictions coincide. For AdamW they are different problems. At matched training loss, AdamW is 29.9 percentage points more accurate than SGD+L2 when the four-coordinate shortcut is reversed at test time, and 22.9 points less accurate when the shortcut is spread over 196 coordinates. Replacing the $\ell_2$-ball projection of plain SGD with the $\ell_\infty$ box that Xie and Li identified for full-batch AdamW reproduces both signs, and a two-feature linear model shows the same reversal exactly. The sign changes near the support at which the hidden units' mean alignment with the shortcut matches its $\ell_2$ value, and Adam with coupled $\ell_2$ shows no reversal that my sweep can resolve. The effect replicates on fresh digit splits and under sign descent with decoupled decay, and fades when the shortcut is much stronger or less reliable.

\end{abstract}

\section{Introduction}
\label{sec:intro}

A spurious shortcut is a feature that predicts the label during training but not at test time, and how much a network relies on it depends on the optimiser. For AdamW it also depends on something that looks irrelevant: how the shortcut is spread across input coordinates. In my setting, a two-layer ReLU network separates the MNIST digit 0 from 1 and 2, and an extra block of inputs carries a shortcut whose sign agrees with the label with probability 0.95. Written on four coordinates, the shortcut resembles a coloured corner patch; spread over 196 coordinates with the same $\ell_2$ norm, it resembles a faint tint. For SGD with coupled $\ell_2$ regularisation (SGD+L2) the two versions are related by an orthogonal change of basis, and over ten seeds SGD+L2 does not change any of its 3{,}920 test predictions between them. AdamW does. At the same training loss, it handles a reversed shortcut much better than SGD+L2 when the shortcut is concentrated and much worse when it is spread out (Figure~\ref{fig:curves}).

The second result runs against the current picture. Networks rely on shortcuts when a simple feature predicts the label well enough~\citep{geirhos2020shortcut,shah2020pitfalls}, and \citet{vasudeva2025rich} found Adam~\citep{kingma2015adam} less prone to this than SGD on a related coloured corner patch, leaving the effect of weight decay, and hence of AdamW, open. \citet{depavia2025rotations} showed on a stylised version of the task that rotating the inputs changes Adam's advantage. \citet{xie2024implicit} give a geometric reason to expect a basis effect from decoupled decay: when full-batch AdamW with decay $\lambda$~\citep{loshchilov2019decoupled} converges under their conditions, it reaches a KKT point of the training loss inside the box $\|W\|_{\max}\le 1/\lambda$. That box is aligned with the coordinate axes, whereas the $\ell_2$ penalty of SGD+L2 is invariant under rotation. The interesting question is whether this geometry explains the reversal, and where it stops explaining it.

I change only the basis in which the shortcut is written, which keeps SGD+L2 provably invariant (Section~\ref{sec:setting}). Keeping the SGD update fixed and replacing its projection onto an $\ell_2$ ball with a projection onto an $\ell_\infty$ box is enough to reproduce the reversal (Section~\ref{sec:verdicts}). I then locate the sign change, test random rotations and other optimisers, and map the conditions under which the reversal disappears (Sections~\ref{sec:sweep}--\ref{sec:travel}). \codeavailability

\section{A shortcut written in two bases}
\label{sec:setting}

\paragraph{Setting.} Each input is $x=(z,\ \alpha s\,u)\in\mathbb{R}^{196}\times\mathbb{R}^{196}$. The digit $z$ is a $14\times14$ MNIST image of a 0 ($y=0$) or of a 1 or 2 ($y=1$), in class-balanced sets of 3{,}000 training, 1{,}000 validation and 1{,}960 test images. In the shortcut block, $u$ is a unit vector and $\alpha=2.5$. On training images $s=2y-1$ with probability $\rho=0.95$ and $s=1-2y$ otherwise. Each test image appears three times, with the shortcut aligned, removed ($s=0$) and reversed ($s=1-2y$). I report accuracy on the reversed copy, a pooled analogue of worst-group accuracy~\citep{sagawa2020distributionally} and the lowest of the three in every run. The direction $u_k=k^{-1/2}\sum_{j\le k}e_j$ spreads the shortcut over $k$ coordinates, so $\|u_k\|_2=1$ while $\|u_k\|_1=\sqrt{k}$. The network is $f_W(x)=\sum_{i=1}^{64}a_i[w_i^\top x]_+$ with a fixed second layer $a_i=\pm1/8$, as in~\citet{vasudeva2025rich}, trained for 1{,}500 steps on mini-batches of 64 with binary cross-entropy (BCE). I compare SGD+L2 ($W\leftarrow W-\eta(\nabla L+\lambda W)$, $\eta=0.1$), Adam+L2 and AdamW with decoupled decay $\lambda$ (both $\eta=10^{-3}$), and plain SGD ($\eta=0.1$) projected after every step onto the ball $\|W\|_F\le R$ or the box $\|W\|_{\max}\le B$, which differ only in the feasible set.

\paragraph{Matching the loss.} The same nominal strength means different things for different optimisers, so I compare them at equal training loss rather than at equal $\lambda$. For each family and seed I walk along a grid of strengths (Appendix~\ref{app:details}) and interpolate every metric linearly at a target training BCE $\tau$. All 600 strength paths of the main sweep have non-decreasing BCE, and the bracketing grid points are at most 0.026 apart in BCE, so the numbers below are interpolated estimates. I use $\tau=0.10$ here (Appendix~\ref{app:tables} adds 0.06 and 0.14), with differences paired by seed and Student-$t$ 95\% intervals over ten seeds.

\begin{lemma}[Transport]
\label{lem:transport}
Let $T\in O(392)$. Replace every input $x$ by $Tx$ and the initialisation $W_0$ by $W_0T^\top$, and keep the mini-batch sequence. Then SGD+L2 and SGD projected onto a Frobenius ball produce the iterates $W_tT^\top$, so $f_{W_tT^\top}(Tx)=f_{W_t}(x)$ for every $x$ and $t$ (proof in Appendix~\ref{app:details}).
\end{lemma}

Every $u_k$ is the image of $u_4$ under a Householder reflection of the shortcut block, so for SGD+L2 all of them are one problem, and float32 agrees up to rounding: in the transport check (one strength per family, ten seeds, five frames) the largest logit change is $3.5\times10^{-5}$ and no test prediction moves, and across the whole sweep a single seed--strength pair of 250 differs between supports, by one test image (Appendix~\ref{app:controls}). Adam+L2, AdamW and the box act coordinate by coordinate, and the lemma says nothing about them.

\paragraph{Why a box depends on the basis.} Consider one hidden unit whose weights lie in a box of radius $B$. By H\"older's inequality, its largest alignment with the shortcut is
\begin{equation}
\label{eq:cap}
\max_{\|w\|_\infty\le B}\, w^\top u \;=\; B\,\|u\|_1 \;=\; B\sqrt{k}\quad\text{for } u=u_k,
\end{equation}
whereas a Frobenius ball allows the same alignment in every direction. A 4-coordinate shortcut therefore reaches the box's limit at $2B$, and a 196-coordinate shortcut at $14B$. Because the box constrains each coordinate separately and the shortcut has its own block, alignment with a dense shortcut uses none of the digit's budget, whereas in a ball the two share one budget. Equation~\eqref{eq:cap} bounds what a unit can represent, not what training reaches, and for AdamW it is an idealisation: Xie and Li assume convergent full-batch iterates, $\epsilon=0$ and a smooth loss, so in my runs $1/\lambda$ is a nominal radius (Appendix~\ref{app:details}).

\section{Results}
\label{sec:results}

\subsection{Does the basis change the verdict?}
\label{sec:verdicts}

\begin{figure}[t]
\centering
\includegraphics[width=\linewidth]{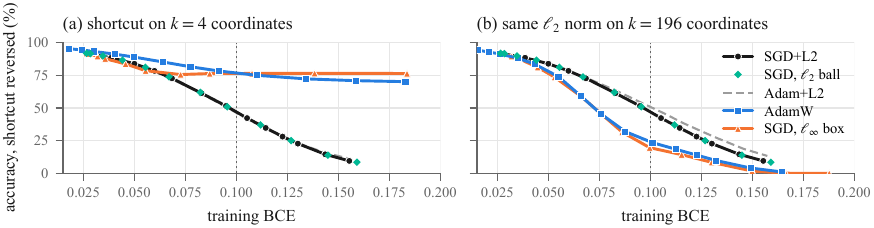}
\caption{Reversed-shortcut accuracy against training BCE as each constraint tightens (ten seeds), for a shortcut on (a)~4 and (b)~196 coordinates with equal $\ell_2$ norm. Dashed: the loss used in Table~\ref{tab:main} and Figure~\ref{fig:sweep}.}
\label{fig:curves}
\end{figure}

Figure~\ref{fig:curves} shows the effect across the whole range of regularisation. The SGD+L2 and ball curves coincide in both panels and show the usual simplicity bias: as the constraint tightens, the network relies more on the shortcut, and reversing it drives accuracy well below chance. With the shortcut on four coordinates, AdamW and the box stay above 70\% across the plotted range; on 196 coordinates, they fall below the SGD+L2 curve.

\begin{table}[tb]
\floatconts{tab:main}{\caption{Interpolated at training BCE 0.10, ten seeds: accuracy (\%) with the shortcut reversed or removed, a shortcut-free last-layer probe (raw features, $C=1$) on reversed inputs, mean alignment $\bar a=\frac1{64}\sum_i|w_i^\top u|$, and paired gaps (pp, 95\% intervals).}}{%
\small
\setlength{\tabcolsep}{4.2pt}
\begin{tabular}{lrrrrrrrr}
\toprule
 & \multicolumn{2}{c}{reversed} & \multicolumn{2}{c}{removed} & \multicolumn{2}{c}{probe, reversed} & \multicolumn{2}{c}{alignment $\bar a$} \\
\cmidrule(lr){2-3}\cmidrule(lr){4-5}\cmidrule(lr){6-7}\cmidrule(lr){8-9}
 & $k=4$ & $k=196$ & $k=4$ & $k=196$ & $k=4$ & $k=196$ & $k=4$ & $k=196$ \\
\midrule
SGD+L2 & 46.9 & 46.9 & 98.2 & 98.2 & 55.2 & 55.2 & 0.084 & 0.084 \\
Adam+L2 & 46.4 & 50.6 & 98.1 & 98.1 & 54.3 & 58.3 & 0.084 & 0.082 \\
SGD, $\ell_2$ ball & 47.1 & 47.1 & 98.2 & 98.2 & 55.2 & 55.2 & 0.084 & 0.084 \\
AdamW & 76.8 & 24.0 & 98.3 & 94.3 & 82.8 & 38.7 & 0.064 & 0.144 \\
SGD, $\ell_\infty$ box & 76.4 & 20.8 & 98.0 & 93.6 & 84.6 & 44.8 & 0.063 & 0.148 \\
\midrule
AdamW $-$ SGD+L2 & \multicolumn{8}{l}{$+29.9$ {\scriptsize[$+26.2$, $+33.6$]} at $k=4$; \quad $-22.9$ {\scriptsize[$-28.3$, $-17.5$]} at $k=196$} \\
box $-$ ball & \multicolumn{8}{l}{$+29.3$ {\scriptsize[$+23.4$, $+35.2$]} at $k=4$; \quad $-26.3$ {\scriptsize[$-29.0$, $-23.7$]} at $k=196$} \\
 
\bottomrule
\end{tabular}}
\end{table}

Table~\ref{tab:main} reports the comparison at a training BCE of 0.10. Moving the shortcut from four coordinates to 196 changes AdamW minus SGD+L2 by $-$52.8~pp [$-$56.3, $-$49.3] and box minus ball by $-$55.6~pp [$-$59.4, $-$51.8]; in the second contrast only the feasible set differs. With the shortcut removed, AdamW loses about four points in the dense basis, so the dense shortcut costs accuracy even when it is absent. The digit itself remains in the hidden features: logistic probes trained on shortcut-free features still decode it at 97.0--98.8\% for every family. On reversed inputs, every decoder I tried keeps AdamW and the box ahead on four coordinates, while on 196 the sign depends on the decoder (Appendix~\ref{app:controls}). The dense shortcut therefore changes how the readout weighs the features more than what the features contain.

Adam with coupled L2 serves as a negative control. Its predictions do change with the basis (Appendix~\ref{app:controls}), but its gap to SGD+L2 changes by +4.2~pp [$-$6.3, +14.8] between the two bases, and all 36 of its paired intervals against SGD+L2 (twelve supports, three losses) contain zero. The intervals leave room for a modest effect; the large reversal appears only with decoupled decay.

\subsection{Where does the sign change?}
\label{sec:sweep}

\begin{figure}[t]
\centering
\includegraphics[width=\linewidth]{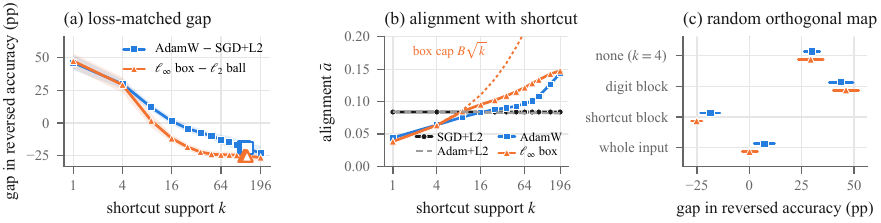}
\caption{Training BCE 0.10, ten seeds. (a)~Paired gaps by support $k$ (pointwise 95\% bands); open markers: random dense directions at $k=\|u\|_1^2$. (b)~Mean alignment $\bar a$ and the box cap $B\sqrt k$; the ball coincides with SGD+L2. (c)~Gaps after Haar-random orthogonal maps from $k=4$ (five draws, bars spanning their 95\% intervals).}
\label{fig:sweep}
\end{figure}

A comparison of feature densities alone would place the sign change near $k\approx69$, where $\sqrt k$ reaches 8.3, the $\ell_1/\ell_2$ ratio of the digit's class-mean difference. The observed change occurs much earlier, between 9 and 25 coordinates for AdamW and between 4 and 16 for the box, judging by where the pointwise 95\% intervals leave zero (Figure~\ref{fig:sweep}a).

Figure~\ref{fig:sweep}b shows what sets the crossing instead. SGD+L2 and the ball settle on the same mean alignment with the shortcut, $\bar a=0.084$, at every $k$, and Adam+L2 stays within 0.002 of it. While the box's cap $B\sqrt k$ lies below that value, the box presses against it (99\% at $k=4$). Once the cap rises past it, the alignment keeps growing, to 43\% of a cap it never reaches at $k=196$, while 39\% of the digit's weights sit within 10\% of the wall. At this loss the gap is indistinguishable from zero near the supports where $\bar a$ matches the $\ell_2$ value, $k=9$ for the box and $k=16$ for AdamW. This is an association at the sampled supports rather than a measured mechanism, since $\bar a$ averages over the first layer and ignores which units the ReLUs switch on. AdamW follows the same pattern less sharply, reaching 57\% of its nominal cap $\sqrt k/\lambda$ at $k=4$ with none of its shortcut weights within 10\% of $1/\lambda$.

\subsection{What do random rotations add?}
\label{sec:rotations}

Figure~\ref{fig:sweep}c applies five Haar-random orthogonal maps to one block, or to the whole input, starting from $u_4$. Rotating the shortcut block gives dense directions with $\|u\|_1$ between 11.12 and 11.56. Interpolating the sweep linearly in $\log k$ at $k=\|u\|_1^2$ gives $-$18.9~pp for AdamW and $-$25.3~pp for box minus ball, and the draws give $-$19.0 to $-$18.4 and $-$25.2 to $-$25.1, so within this range $\|u\|_1$ predicts the gap to within about half a point. Rotating the digit block has the opposite effect: the digit becomes denser and both gaps widen.

Rotating the whole input mixes the digit and shortcut coordinates, and the box no longer explains AdamW. Box minus ball is +0.1~pp [$-$1.9, +2.0] when the five draws are pooled within each seed, whereas AdamW keeps +5.8 to +8.4~pp over SGD+L2 and +4.0~pp [+0.3, +7.8] over Adam+L2. This difference is exploratory, measured on five maps.

\subsection{How far does the reversal travel?}
\label{sec:travel}

Two controls come first. With an uninformative shortcut ($\rho=0.5$) the large reversal vanishes (97.2--99.4\% reversed accuracy for every family), and the AdamW gap keeps its sign at each support when I train the second layer and biases, halve or double the step sizes, or double the number of steps. Three fresh digit splits reproduce the reversal, and the worst of the four label-by-shortcut groups moves the same way; so does a weaker shortcut ($\alpha=1.25$). Sign descent with decoupled decay and AdamW with a cosine schedule show it as well, while SGD with momentum and coupled L2 stays basis-invariant (Appendix~\ref{app:controls}). The reversal also has clear limits. With a shortcut twice as strong ($\alpha=5$) the dense-basis penalty disappears. When the shortcut agrees with the label only 90\% or 80\% of the time, the gaps at BCE 0.10 shrink to at most 9 points, because at that loss the network barely needs the shortcut; matched instead at half the loss of a shortcut-only predictor, the dense penalty returns at both levels and the sparse advantage only at 90\%. In a two-feature linear model the geometric effect can be computed exactly (Appendix~\ref{app:linear}): at equal loss, the box beats the ball when the shortcut is sparser than the core and falls to chance when both are equally spread.

\section{Related work and conclusion}
\label{sec:conclusion}

Optimisation geometry shapes implicit bias~\citep{gunasekar2018characterizing,wilson2017marginal}, sign-based and adaptive methods follow $\ell_\infty$ geometry~\citep{bernstein2018signsgd,balles2018dissecting,xie2025adam,zhang2024implicit,xie2024implicit}, and the basis affects Adam's speed~\citep{zhang2025understanding} and feature richness~\citep{depavia2025rotations}. \citet{ray2026amplifies} found that a synthetic shortcut's logit sensitivity grew less with dataset size under Adam and AdamW than under unregularised SGD. My contribution is the regularised case, a matched-loss support sweep in which the basis sets the sign of AdamW's effect, with a projection control that isolates the constraint set. The setting is small (one task, 64 units, a shortcut in its own block), and the effect depends on the shortcut's strength and reliability. Real shortcuts overlap with the object and deep networks re-encode their inputs, so for AdamW the relevant coordinates are each layer's parameters rather than the pixels.

Within these limits, a corner patch and a tint of equal energy are the same problem for SGD+L2 and two different problems for AdamW, depending on how the shortcut is spread relative to the digit: coupled $\ell_2$ regularisation is invariant to that distribution, and decoupled weight decay is not.

\bibliographystyle{plainnat}
\bibliography{references}

\clearpage
\appendix
\section{Experimental details}
\label{app:details}

\paragraph{Proof of Lemma~\ref{lem:transport}.} Since $WT^\top Tx=Wx$, the mini-batch loss at $WT^\top$ on the transformed inputs equals the loss at $W$ on the original ones, and its gradient is $\nabla L(W)\,T^\top$; at the kinks of the ReLU the same derivative convention applies to equal pre-activations. The coupled update $W-\eta(\nabla L+\lambda W)$ is linear in $(W,\nabla L)$ and commutes with right-multiplication by $T^\top$. The projection rescales by $\min(1,R/\|W\|_F)$, and $\|WT^\top\|_F=\|W\|_F$. Induction on $t$ completes the proof. The argument holds in exact arithmetic; Appendix~\ref{app:controls} measures the float32 deviations.

\paragraph{Data.} MNIST images of the digits 0, 1 and 2 are resized to $14\times14$ with antialiasing (torchvision) and kept in $[0,1]$. A single class-balanced split, drawn once with a fixed generator, gives 3{,}000 training and 1{,}000 validation images from the official training set and 1{,}960 test images from the official test set. The seed changes the shortcut signs of the training images, the initialisation, the fixed second layer and the mini-batch order, never the split.

\paragraph{Initialisation and transport.} In the reference frame ($k=4$) the first layer starts from independent $\mathcal N(0,(10^{-3}/\sqrt{392})^2)$ entries, as in~\citet{vasudeva2025rich}. Every other frame uses $W_0T^\top$, where $T$ is the orthogonal map that takes the reference inputs to that frame: a Householder reflection of the shortcut block for $u_k$, or a Haar-random orthogonal matrix of one block or of the whole input in Section~\ref{sec:rotations}. Mini-batch indices are drawn with replacement from a seed-specific generator and are identical across families, strengths and frames.

\paragraph{Strength grids.} For SGD+L2 and Adam+L2 I use $\lambda$ = 0, 0.005, 0.01, 0.015, 0.02, 0.03, 0.04, 0.05, 0.06, 0.07, 0.08, 0.10, 0.12. For AdamW I use $\lambda$ = 0, 1, 2, 3, 5, 7, 10, 13, 16, 20, 25, 30, 35, 40, 50, 60. For the ball I use $R$ = 5, 4, 3.5, 3, 2.6, 2.3, 2, 1.8, 1.6, 1.45, 1.3, 1.2. For the box I use $B$ = 0.2, 0.15, 0.1, 0.08, 0.06, 0.05, 0.04, 0.035, 0.03, 0.025, 0.02, 0.017, 0.014, 0.012, 0.01. Adam and AdamW use the PyTorch defaults $\beta=(0.9,0.999)$ and $\epsilon=10^{-8}$. PyTorch's AdamW multiplies the weights by $1-\eta\lambda$ before each Adam step. \citet{xie2024implicit} show that full-batch AdamW iterates with $\epsilon=0$, $\beta_1\le\beta_2<1$, non-increasing steps whose sum diverges and a continuously differentiable loss reach, if they converge, a KKT point of the loss inside the box of radius $1/\lambda$. My runs use mini-batches, $\epsilon>0$ and ReLU units, so I treat $1/\lambda$ as a nominal radius. In fact, 111 of the 1{,}800 AdamW endpoints with $\lambda>0$ in the sweep exceed it somewhere in $W$, by at most 26\%.

\paragraph{Metrics.} Reversed, removed and aligned accuracies are computed on the same 1{,}960 test digits. The alignment is $\bar a=\frac1{64}\sum_i|w_i^\top u|$ for the trained first layer and the shortcut direction $u$ of the frame. The cap fraction is $\bar a/(B\|u\|_1)$, with $B$ the box radius or $1/\lambda$ for AdamW. A weight counts as within 10\% of the wall if $|w_{ij}|\ge0.9B$. The probe is a logistic regression (scikit-learn, $C=1$) on the 64 hidden activations of the 1{,}000 validation images with the shortcut removed, evaluated on the reversed test images. For the interpolation at a target loss $\tau$, each path is ordered from the weakest to the strongest constraint, and the first segment whose two training losses bracket $\tau$ is used. Every per-run metric, ratios included, is interpolated within its seed first and averaged over seeds afterwards; at $\tau=0.10$ the bracketing points are at most 0.026 apart in BCE (median 0.015). All intervals are conditional on the single digit split.

\paragraph{Runs and compute.} The result files hold one row per run: 8{,}280 for the sweep, 8{,}550 for the random orthogonal maps, 300 for the negative control, 840 for the trainable second layer, 1{,}320 for the step-size and horizon check, 20 for the replication, 4{,}760 for the confirmation and boundary runs, 980 for the other optimisers and 160 for the probe check, plus 250 comparisons in the transport check, which trains 300 networks. One pass of the suite therefore trains 25{,}510 networks. Each runs on CPU with one thread per process, which makes it deterministic and independent of the number of workers; the recorded stage timings add up to 15 minutes with 16 workers on a laptop. Re-running the random-map experiment reproduced its earlier 6{,}600 rows exactly. A script regenerates every table from the per-run files and checks a list of the numbers quoted in the text, each in the file where it appears, without retraining.

\section{Full tables}
\label{app:tables}

Tables~\ref{tab:sweep} and~\ref{tab:sweep_adam} give the paired gaps behind Figure~\ref{fig:sweep}a at three matched losses, and Table~\ref{tab:rotations} the per-draw results behind Figure~\ref{fig:sweep}c.

\begin{table}[!ht]
\floatconts{tab:sweep}{\caption{Paired gaps in reversed accuracy (pp, mean $\pm$ half-width of the 95\% interval, ten seeds) against the shortcut's support $k$ (so $\|u_k\|_1=\sqrt k$), at three matched training losses. SGD+L2 and the ball are identical across $k$, so every change down a column comes from AdamW or the box.}}{%
\small
\setlength{\tabcolsep}{4pt}
\begin{tabular}{rrrrrrr}
\toprule
 & \multicolumn{3}{c}{AdamW $-$ SGD+L2} & \multicolumn{3}{c}{box $-$ ball} \\
\cmidrule(lr){2-4}\cmidrule(lr){5-7}
$k$ & 0.06 & 0.10 & 0.14 & 0.06 & 0.10 & 0.14 \\
\midrule
1 & $+15.7\pm3.6$ & $+46.4\pm5.7$ & $+76.1\pm4.3$ & $+16.3\pm4.0$ & $+47.5\pm5.7$ & $+77.3\pm4.7$ \\
4 & $+8.1\pm1.7$ & $+29.9\pm3.7$ & $+54.8\pm3.0$ & $+0.2\pm3.0$ & $+29.3\pm5.9$ & $+59.4\pm4.9$ \\
9 & $+2.8\pm1.3$ & $+12.0\pm2.6$ & $+27.2\pm2.8$ & $-7.8\pm1.4$ & $+1.8\pm5.1$ & $+30.1\pm4.5$ \\
16 & $-0.4\pm2.1$ & $+1.5\pm2.7$ & $+9.7\pm3.2$ & $-11.2\pm1.0$ & $-11.7\pm2.9$ & $+7.4\pm3.6$ \\
25 & $-2.1\pm2.8$ & $-4.2\pm2.9$ & $+0.5\pm3.6$ & $-12.4\pm1.3$ & $-18.3\pm2.4$ & $-5.6\pm3.4$ \\
36 & $-2.9\pm2.9$ & $-7.2\pm2.6$ & $-4.1\pm3.8$ & $-12.5\pm1.4$ & $-22.1\pm2.4$ & $-10.3\pm3.0$ \\
49 & $-3.3\pm2.7$ & $-9.8\pm2.5$ & $-7.2\pm3.3$ & $-12.5\pm1.4$ & $-23.7\pm2.3$ & $-12.1\pm3.0$ \\
64 & $-3.5\pm2.5$ & $-12.5\pm3.6$ & $-8.5\pm3.4$ & $-12.5\pm1.4$ & $-24.3\pm2.5$ & $-12.4\pm3.2$ \\
81 & $-3.6\pm2.7$ & $-14.1\pm4.1$ & $-8.8\pm3.5$ & $-12.5\pm1.4$ & $-24.3\pm2.7$ & $-11.5\pm3.0$ \\
100 & $-4.1\pm2.9$ & $-16.3\pm4.2$ & $-9.0\pm3.7$ & $-12.5\pm1.4$ & $-24.6\pm2.6$ & $-10.5\pm2.9$ \\
144 & $-6.6\pm5.1$ & $-20.2\pm4.4$ & $-9.5\pm4.0$ & $-12.5\pm1.4$ & $-25.6\pm2.6$ & $-10.8\pm2.8$ \\
196 & $-10.3\pm6.0$ & $-22.9\pm5.4$ & $-10.2\pm4.1$ & $-12.5\pm1.4$ & $-26.3\pm2.7$ & $-12.0\pm3.0$ \\
 
\bottomrule
\end{tabular}}
\end{table}

\begin{table}[!ht]
\floatconts{tab:sweep_adam}{\caption{Adam+L2 minus SGD+L2, same format as Table~\ref{tab:sweep}. All 36 intervals contain zero.}}{%
\small
\begin{tabular}{rrrr}
\toprule
$k$ & 0.06 & 0.10 & 0.14 \\
\midrule
1 & $-0.1\pm0.9$ & $-0.1\pm1.5$ & $+1.2\pm2.5$ \\
4 & $-0.5\pm1.6$ & $-0.5\pm2.9$ & $+0.8\pm3.2$ \\
9 & $-0.9\pm2.2$ & $-0.5\pm3.8$ & $+1.2\pm3.7$ \\
16 & $-0.9\pm2.6$ & $+0.4\pm3.8$ & $+2.0\pm4.1$ \\
25 & $-0.4\pm2.5$ & $+1.4\pm3.7$ & $+2.6\pm4.5$ \\
36 & $+0.1\pm2.1$ & $+2.3\pm4.1$ & $+2.9\pm4.9$ \\
49 & $+0.4\pm1.8$ & $+2.8\pm4.8$ & $+3.5\pm5.7$ \\
64 & $+0.6\pm1.8$ & $+2.7\pm5.3$ & $+3.9\pm6.4$ \\
81 & $+0.5\pm2.3$ & $+2.8\pm6.1$ & $+4.4\pm7.2$ \\
100 & $+0.4\pm2.6$ & $+2.9\pm6.9$ & $+5.0\pm7.9$ \\
144 & $+0.1\pm3.4$ & $+3.2\pm8.2$ & $+6.3\pm8.9$ \\
196 & $-0.2\pm3.8$ & $+3.8\pm8.7$ & $+6.1\pm8.3$ \\
 
\bottomrule
\end{tabular}}
\end{table}

\begin{table}[!ht]
\floatconts{tab:rotations}{\caption{Random orthogonal maps applied to one block, or to the whole input, starting from $k=4$ (matched training BCE 0.10, ten seeds per draw; pp with 95\% intervals). $\|u\|_1$ is the shortcut's $\ell_1$ norm and $\kappa_z$ the $\ell_1/\ell_2$ ratio of the digit's class-mean difference after the map; for the whole input both are computed in $\mathbb{R}^{392}$. In the pixel basis $\kappa_z=8.30$. The ball is invariant (Lemma~\ref{lem:transport}), so its runs from the reference frame serve every draw.}}{%
\small
\begin{tabular}{lcrrll}
\toprule
map applied to & draw & $\|u\|_1$ & $\kappa_z$ & AdamW $-$ SGD+L2 & box $-$ ball \\
\midrule
digit block & 1 & 2.00 & 11.13 & $+43.8$ {\scriptsize[$+38.5$, $+49.1$]} & $+46.0$ {\scriptsize[$+40.3$, $+51.7$]} \\
 & 2 & 2.00 & 10.92 & $+43.5$ {\scriptsize[$+38.2$, $+48.9$]} & $+45.8$ {\scriptsize[$+40.1$, $+51.5$]} \\
 & 3 & 2.00 & 10.96 & $+43.5$ {\scriptsize[$+38.2$, $+48.8$]} & $+46.2$ {\scriptsize[$+40.4$, $+51.9$]} \\
 & 4 & 2.00 & 11.41 & $+44.2$ {\scriptsize[$+38.9$, $+49.5$]} & $+46.5$ {\scriptsize[$+40.8$, $+52.2$]} \\
 & 5 & 2.00 & 11.29 & $+43.7$ {\scriptsize[$+38.3$, $+49.0$]} & $+46.2$ {\scriptsize[$+40.5$, $+51.9$]} \\
\addlinespace
shortcut block & 1 & 11.21 & 8.30 & $-18.7$ {\scriptsize[$-23.0$, $-14.4$]} & $-25.1$ {\scriptsize[$-27.6$, $-22.7$]} \\
 & 2 & 11.12 & 8.30 & $-18.4$ {\scriptsize[$-22.7$, $-14.0$]} & $-25.1$ {\scriptsize[$-27.6$, $-22.6$]} \\
 & 3 & 11.13 & 8.30 & $-18.4$ {\scriptsize[$-22.7$, $-14.0$]} & $-25.1$ {\scriptsize[$-27.6$, $-22.6$]} \\
 & 4 & 11.32 & 8.30 & $-18.9$ {\scriptsize[$-23.3$, $-14.6$]} & $-25.1$ {\scriptsize[$-27.6$, $-22.7$]} \\
 & 5 & 11.56 & 8.30 & $-19.0$ {\scriptsize[$-23.3$, $-14.7$]} & $-25.2$ {\scriptsize[$-27.7$, $-22.7$]} \\
\addlinespace
whole input & 1 & 16.00 & 15.86 & $+7.2$ {\scriptsize[$+3.2$, $+11.1$]} & $+0.2$ {\scriptsize[$-1.7$, $+2.0$]} \\
 & 2 & 15.58 & 15.71 & $+5.8$ {\scriptsize[$+2.3$, $+9.2$]} & $-0.2$ {\scriptsize[$-2.1$, $+1.6$]} \\
 & 3 & 16.05 & 15.81 & $+7.8$ {\scriptsize[$+3.8$, $+11.8$]} & $-0.3$ {\scriptsize[$-2.5$, $+1.9$]} \\
 & 4 & 15.73 & 15.67 & $+8.4$ {\scriptsize[$+4.7$, $+12.0$]} & $+1.9$ {\scriptsize[$-0.1$, $+3.9$]} \\
 & 5 & 15.61 & 15.87 & $+7.1$ {\scriptsize[$+3.2$, $+11.0$]} & $-1.2$ {\scriptsize[$-3.4$, $+1.0$]} \\
 
\bottomrule
\end{tabular}}
\end{table}

\FloatBarrier
\section{Controls}
\label{app:controls}

\paragraph{Transport check.} For ten seeds, each family was trained at one strength (SGD+L2 and Adam+L2 with $\lambda=0.05$, AdamW with $\lambda=20$, ball $R=2$, box $B=0.03$) in the reference frame and in five others ($k=1$, $k=196$, and one random map of the shortcut block, of the digit block and of the whole input), and the 3{,}920 test logits of each comparison (aligned and reversed copies of the 1{,}960 test images) were compared. For SGD+L2 and the ball, the largest absolute logit difference from the reference frame is $1.6\times10^{-5}$ and $3.5\times10^{-5}$ (logits reach 5.7 in magnitude), the transported weights agree to $2.1\times10^{-5}$, and no prediction changes. Adam+L2 changes between 13 and 461 of the 3{,}920 predictions per run, AdamW between 178 and 1{,}027, and the box between 247 and 1{,}027. In the main sweep, the SGD+L2 and ball runs at a given seed and strength reach the same training BCE at every support to within $2.6\times10^{-6}$, and the same reversed accuracy in all 250 seed--strength pairs but one (SGD+L2, seed 6, $\lambda=0.10$), where one of the 1{,}960 reversed test images flips.

\paragraph{Negative control.} With $\rho=0.5$ the shortcut sign is independent of the label. Table~\ref{tab:negative} lists, for each family, the range over three strengths of ten-seed means. Individual runs spread a little wider, from 96.5\% to 99.7\% reversed accuracy, and the largest share of changed predictions in a single run is 2.1\% (42 of 1{,}960 images, AdamW, $k=196$, $\lambda=30$).

\begin{table}[!ht]
\floatconts{tab:negative}{\caption{Negative control, $\rho=0.5$. For each family, the range over three strengths of the ten-seed mean accuracy (\%) with the shortcut reversed and removed, and the largest ten-seed mean share (\%) of test predictions that change when the shortcut is reversed.}}{%
\small
\begin{tabular}{lrrrrrr}
\toprule
 & \multicolumn{3}{c}{$k=4$} & \multicolumn{3}{c}{$k=196$} \\
\cmidrule(lr){2-4}\cmidrule(lr){5-7}
 & reversed & removed & changed & reversed & removed & changed \\
\midrule
SGD+L2 & 98.4--99.3 & 98.3--99.3 & 0.3 & 98.4--99.3 & 98.3--99.3 & 0.3 \\
Adam+L2 & 98.3--99.4 & 98.2--99.4 & 0.4 & 98.2--99.3 & 98.2--98.7 & 0.8 \\
SGD, $\ell_2$ ball & 98.2--99.2 & 98.1--99.2 & 0.3 & 98.2--99.2 & 98.1--99.2 & 0.3 \\
AdamW & 98.0--99.4 & 97.8--99.4 & 0.4 & 97.8--99.3 & 92.9--98.7 & 0.9 \\
SGD, $\ell_\infty$ box & 97.2--99.2 & 97.0--99.2 & 0.7 & 97.2--99.2 & 97.0--99.2 & 0.7 \\
 
\bottomrule
\end{tabular}}
\end{table}

In the dense basis, AdamW at $\lambda=30$ loses accuracy when the shortcut is removed (92.9\% on average) while keeping 97.8\% when it is reversed. Removing the shortcut puts the input off the training distribution, where every training image carried $s=\pm1$.

\paragraph{Trainable second layer, step size and horizon.} Table~\ref{tab:head} repeats the main contrast with both layers and the biases trained, all parameters regularised. Table~\ref{tab:controls} halves or doubles every step size, or doubles the number of steps, with five seeds; each family is compared with SGD+L2 run under the same change.

\begin{table}[!ht]
\floatconts{tab:head}{\caption{Second layer and biases trained (pp, mean $\pm$ half-width of the 95\% interval, ten seeds).}}{%
\small
\begin{tabular}{crrrr}
\toprule
 & \multicolumn{2}{c}{AdamW $-$ SGD+L2} & \multicolumn{2}{c}{Adam+L2 $-$ SGD+L2} \\
\cmidrule(lr){2-3}\cmidrule(lr){4-5}
training BCE & $k=4$ & $k=196$ & $k=4$ & $k=196$ \\
\midrule
0.06 & $+9.6\pm2.6$ & $-9.9\pm5.3$ & $-0.7\pm2.0$ & $-0.1\pm3.8$ \\
0.10 & $+33.2\pm4.3$ & $-20.8\pm3.1$ & $-0.5\pm3.7$ & $+1.5\pm7.1$ \\
 
\bottomrule
\end{tabular}}
\end{table}

\begin{table}[!ht]
\floatconts{tab:controls}{\caption{Step size and horizon at matched training BCE 0.10 (pp, mean $\pm$ half-width of the 95\% interval, five seeds).}}{%
\small
\begin{tabular}{lrrrr}
\toprule
 & \multicolumn{2}{c}{AdamW $-$ SGD+L2} & \multicolumn{2}{c}{box $-$ SGD+L2} \\
\cmidrule(lr){2-3}\cmidrule(lr){4-5}
change & $k=4$ & $k=196$ & $k=4$ & $k=196$ \\
\midrule
step sizes $\times 0.5$ & $+35.4\pm6.4$ & $-20.8\pm2.9$ & $+33.0\pm10.1$ & $-25.2\pm4.8$ \\
step sizes $\times 2$ & $+31.7\pm5.3$ & $-20.9\pm11.3$ & $+34.5\pm12.7$ & $-24.1\pm7.6$ \\
3{,}000 steps & $+29.2\pm5.0$ & $-18.3\pm8.0$ & $+32.9\pm6.3$ & $-25.7\pm4.7$ \\
 
\bottomrule
\end{tabular}}
\end{table}

\paragraph{Shared strength.} Matching the loss tunes each family separately. At a shared decay of $\lambda=20$, AdamW's reversed accuracy is 75.0\% on four coordinates and 31.9\% on 196, at training BCE 0.110 and 0.088; without decay it is 96.2\% and 95.7\%.

\paragraph{Confirmation and boundaries.} Table~\ref{tab:confirm} repeats the endpoint contrasts with five seeds on three fresh class-balanced splits (drawn with generator seeds 101, 202 and 303 from the same MNIST pools), with the shortcut amplitude halved or doubled, and with the shortcut agreeing with the label 90\% or 80\% of the time. The strength grids are extended towards stronger constraints so that every path brackets the targets. For $\rho<0.95$ I also match at half the BCE of a predictor that uses only the shortcut, $\tfrac12 H(\rho)$, which rounds to 0.10 at $\rho=0.95$. On the fresh splits, the worst of the four label-by-shortcut groups gives AdamW gaps of +34.6 to +36.4~pp on four coordinates and $-$31.9 to $-$29.7~pp on 196.

\begin{table}[!ht]
\floatconts{tab:confirm}{\caption{Confirmation and boundary runs (pp, mean $\pm$ half-width of the 95\% interval, five seeds, reversed accuracy). $\tau$ is the matched training BCE.}}{%
\small
\begin{tabular}{lcrrrr}
\toprule
 & & \multicolumn{2}{c}{AdamW $-$ SGD+L2} & \multicolumn{2}{c}{box $-$ ball} \\
\cmidrule(lr){3-4}\cmidrule(lr){5-6}
condition & $\tau$ & $k=4$ & $k=196$ & $k=4$ & $k=196$ \\
\midrule
fresh split 1 & 0.10 & $+31.9\pm4.6$ & $-21.3\pm13.6$ & $+31.8\pm8.7$ & $-24.6\pm6.4$ \\
fresh split 2 & 0.10 & $+32.3\pm5.5$ & $-21.5\pm12.4$ & $+32.7\pm11.6$ & $-24.4\pm7.3$ \\
fresh split 3 & 0.10 & $+31.0\pm4.9$ & $-18.5\pm14.8$ & $+32.6\pm10.5$ & $-22.6\pm8.2$ \\
$\alpha=1.25$ & 0.10 & $+13.0\pm2.3$ & $-44.9\pm8.0$ & $+14.4\pm2.6$ & $-60.7\pm4.1$ \\
$\alpha=5$ & 0.10 & $+29.8\pm7.2$ & $+1.8\pm6.8$ & $+16.8\pm3.9$ & $+0.9\pm3.5$ \\
$\rho=0.90$ & 0.10 & $+2.3\pm1.4$ & $-8.8\pm7.1$ & $-2.1\pm1.9$ & $-7.1\pm1.4$ \\
$\rho=0.90$ & 0.16 & $+12.2\pm2.9$ & $-35.8\pm7.6$ & $+10.1\pm4.8$ & $-32.8\pm6.1$ \\
$\rho=0.80$ & 0.10 & $-0.1\pm0.2$ & $-1.4\pm0.8$ & $-1.0\pm0.2$ & $-1.0\pm0.2$ \\
$\rho=0.80$ & 0.25 & $+0.4\pm1.0$ & $-36.3\pm19.2$ & $-5.5\pm2.1$ & $-12.3\pm1.7$ \\
 
\bottomrule
\end{tabular}}
\end{table}

\paragraph{Other optimisers.} Table~\ref{tab:optimisers} compares three more update rules with SGD+L2 on the main split, with ten seeds and the same pairing: SGD with heavy-ball momentum 0.9, step 0.01 and coupled L2; sign descent with decoupled decay, $W\leftarrow(1-\eta\lambda)W-\eta\,\mathrm{sign}(\nabla L)$ with $\eta=10^{-3}$; and AdamW whose step size follows a cosine schedule from $10^{-3}$ to zero. Momentum SGD gives identical reversed accuracy at the two supports for every seed and strength, as transport predicts. The sign-descent paths are non-monotone only below a training BCE of 0.03, far from the matched losses.

\begin{table}[!ht]
\floatconts{tab:optimisers}{\caption{Other update rules minus SGD+L2 (pp, mean $\pm$ half-width of the 95\% interval, ten seeds, reversed accuracy at three matched training losses).}}{%
\small
\setlength{\tabcolsep}{3pt}
\begin{tabular}{lrrrrrr}
\toprule
 & \multicolumn{2}{c}{$\tau=0.06$} & \multicolumn{2}{c}{$\tau=0.10$} & \multicolumn{2}{c}{$\tau=0.14$} \\
\cmidrule(lr){2-3}\cmidrule(lr){4-5}\cmidrule(lr){6-7}
 & $k=4$ & $k=196$ & $k=4$ & $k=196$ & $k=4$ & $k=196$ \\
\midrule
momentum SGD+L2 & $+0.1\pm0.3$ & $+0.1\pm0.3$ & $+0.1\pm1.0$ & $+0.1\pm1.0$ & $-0.0\pm1.4$ & $-0.0\pm1.4$ \\
sign descent & $+5.5\pm1.9$ & $-7.1\pm6.4$ & $+24.4\pm5.2$ & $-24.2\pm6.5$ & $+50.8\pm5.0$ & $-13.0\pm4.2$ \\
AdamW, cosine & $+8.7\pm2.7$ & $-9.9\pm3.8$ & $+30.1\pm5.1$ & $-23.9\pm4.1$ & $+55.2\pm4.3$ & $-11.8\pm3.3$ \\
AdamW, constant & $+8.1\pm1.7$ & $-10.3\pm6.0$ & $+29.9\pm3.7$ & $-22.9\pm5.4$ & $+54.8\pm3.0$ & $-10.2\pm4.1$ \\
 
\bottomrule
\end{tabular}}
\end{table}

\paragraph{Probes.} For the 160 runs that bracket training BCE 0.10 at $k=4$ and $k=196$ (four families, ten seeds), I retrained the network and fitted six logistic decoders on the hidden activations of the shortcut-free validation images, with $C\in\{0.01,1,100\}$ on raw or standardised features. Table~\ref{tab:probes} gives their accuracy on shortcut-free and on reversed test images, interpolated at the matched loss. The column of Table~\ref{tab:main} is the raw, $C=1$ row.

\begin{table}[!ht]
\floatconts{tab:probes}{\caption{Last-layer probes at matched training BCE 0.10 (ten seeds). Range over the four families of the accuracy on shortcut-free test images (\%), and paired gaps on reversed test images (pp, mean $\pm$ half-width of the 95\% interval).}}{%
\small
\setlength{\tabcolsep}{4pt}
\begin{tabular}{lcrrrr}
\toprule
 & & \multicolumn{2}{c}{AdamW $-$ SGD+L2} & \multicolumn{2}{c}{box $-$ ball} \\
\cmidrule(lr){3-4}\cmidrule(lr){5-6}
decoder & removed & $k=4$ & $k=196$ & $k=4$ & $k=196$ \\
\midrule
raw, $C=0.01$ & 97.0--98.4 & $+24.8\pm3.8$ & $-2.1\pm4.0$ & $+26.0\pm5.5$ & $-7.7\pm2.9$ \\
raw, $C=1$ & 97.3--98.6 & $+27.6\pm4.3$ & $-16.4\pm9.1$ & $+29.4\pm6.2$ & $-10.4\pm9.0$ \\
raw, $C=100$ & 98.0--98.6 & $+27.2\pm4.3$ & $-5.7\pm24.6$ & $+29.1\pm6.3$ & $+21.0\pm20.0$ \\
standardised, $C=0.01$ & 97.2--98.5 & $+30.6\pm4.1$ & $-32.2\pm8.5$ & $+33.6\pm6.1$ & $-20.9\pm6.4$ \\
standardised, $C=1$ & 97.8--98.6 & $+27.4\pm4.3$ & $-6.4\pm23.9$ & $+29.3\pm6.3$ & $+20.5\pm18.8$ \\
standardised, $C=100$ & 98.2--98.8 & $+23.7\pm4.1$ & $+1.5\pm23.7$ & $+29.0\pm6.3$ & $+25.1\pm15.5$ \\
 
\bottomrule
\end{tabular}}
\end{table}

\paragraph{Replication of the corner patch.} I follow Section~4.2 and Appendix~B.2 of~\citet{vasudeva2025rich}: $14\times14$ RGB MNIST, digit 0 against 1 and 2, a $2\times2$ top-left patch that is red for one class and green for the other ($\rho=1$), width 64, fixed second layer $\pm1/8$, initialisation scale $10^{-3}/\sqrt d$, SGD with step size 0.1 and Adam with $10^{-3}$, no regularisation, batches of 64 and a stopping BCE of 0.005. Implementation choices not fixed by their description are mine: mini-batches drawn with replacement, the stopping loss evaluated on the full training set every 50 steps, and a cap of 4{,}000 steps. On the full training set (18{,}623 images) and test set (3{,}147), with the patch colours swapped at test time, five seeds give $66.2\pm1.6\%$ reversed-patch accuracy for SGD and $90.2\pm1.1\%$ for Adam (mean $\pm$ standard deviation), against $66.6\%$ and $86.0\%$ in their paper. The ordering and the SGD figure reproduce; my Adam figure is 4.2 points higher. SGD reaches the stopping loss after $1{,}360\pm114$ steps and Adam after $490\pm42$. On a class-balanced subset of 6{,}000 training and 1{,}960 test images the figures are $61.5\%$ and $90.2\%$.

\section{Same loss, opposite verdicts in a linear model}
\label{app:linear}

The trained networks mix geometry with training dynamics. A two-feature linear model isolates the constraint by comparing its exact optima at the same population loss.

\begin{proposition}
\label{prop:linear}
Let $y\in\{-1,+1\}$ be equiprobable and $x=y\,(Zv+Su)$, where $v$ and $u$ are unit vectors with disjoint supports and $Z\sim\mathcal N(1,0.1)$ and $S\in\{-1,+1\}$ with $\mathbb E S=0.8$ are independent. For a linear predictor $w^\top x$ write $a=w^\top v$ and $b=w^\top u$. Take $v$ uniform on four coordinates, so $\|v\|_1=2$.
On the ball $\|w\|_2\le 0.5596$, whatever the unit vector $u$, the square loss is minimised at $(a,b)=(0.4886,0.2727)$, with loss $L^\star=0.1366$ and accuracy 91.9\% on reversed inputs ($S=-1$).

On the box $\|w\|_\infty\le B$, with $u$ on a single coordinate and $B=0.2507$, the minimiser is $(a,b)=(0.5014,0.2507)$, with loss $L^\star$ and reversed accuracy 94.3\%. With $u$ uniform on four coordinates and $B=0.2144$, it is $(0.4289,0.4289)$, again with loss $L^\star$, and reversed accuracy 50.0\%.
\end{proposition}
\begin{proof}
The prediction depends on $w$ only through $(a,b)$, and $\mathbb E(y-w^\top x)^2=\mathbb E(1-aZ-bS)^2=(1-a-0.8b)^2+0.1a^2+0.36b^2=L(a,b)$. On a reversed input $y\,w^\top x=aZ-b$, so for $a>0$ the accuracy is $\Pr[Z>b/a]$. For orthonormal $v,u$ the ball constrains $a^2+b^2\le R^2$, whatever $u$ is. The stated point solves $\nabla L(a,b)+(a,b)=0$, so it minimises $L$ on the disc of its own radius, because the unconstrained minimiser of $L$, $(0.7826,0.1739)$, lies outside that disc. With disjoint supports the box allows exactly the rectangle $|a|\le B\|v\|_1$, $|b|\le B\|u\|_1$. At each stated corner both partial derivatives of $L$ are negative, so the corner satisfies the KKT conditions of this convex problem and minimises $L$ on the rectangle. The radius $B$ is the smaller root of $L(2B,B\|u\|_1)=L^\star$. The accuracies follow from $Z\sim\mathcal N(1,0.1)$. The numbers are computed exactly, and checked by brute force on a $2001\times2001$ grid, in \texttt{experiments/linear\_example.py}.
\end{proof}

The two box problems differ only by an orthogonal change of basis inside the shortcut's block, which leaves the ball problem unchanged. In this example the box's accuracy on reversed inputs depends on the shortcut's support even at equal loss.

\end{document}